\documentclass[]{fairmeta}
\usepackage{amsmath}
\usepackage{amssymb}
\usepackage{booktabs}
\usepackage{graphicx}
\usepackage[table]{xcolor}
\usepackage{adjustbox}
\usepackage{multirow}
\usepackage{microtype}
\usepackage{hyperref}
\usepackage{url}
\usepackage{wrapfig}
\usepackage{subcaption}
\usepackage{caption} 

\title{SGA-Sparse: Training Native SwiGLU Gates for Activation-Sparse and Low-Bit LLMs}

\newcommand{\bn}[1]{\mathbf{#1}}
\newcommand{\name}{\textbf{SGA-Sparse}~}
\definecolor{sakura}{cmyk}{0,0.15,0.05,0.05}

\author{Ruokai Yin}
\author{Steven Li}

\affiliation{Meta AI}

\abstract{Activation sparsity can reduce FFN weight traffic during single-user LLM decoding, but active channels must be identified early enough to avoid fetching unused weights. In pretrained SwiGLU models, training-free routing with the native gate performs substantially worse than routing with the up activation at high sparsity. We show that this ordering reverses after sparsity-aware training. Based on this insight, we introduce \textbf{SGA-Sparse}, which converts pretrained SwiGLU LLMs into fixed-budget activation-sparse models by training the existing gate projection directly as a top-$k$ router. \textbf{SGA-Sparse} updates all model parameters end-to-end while preserving the original SwiGLU activation and introducing no auxiliary router parameters. After approximately 30B training tokens, LLaMA models from 1B to 8B achieve uniform per-token 75--85\% FFN channel sparsity while remaining within 0.2--2.7 pp of their original dense checkpoints in average zero-shot accuracy. Because post-training quantization degrades sparsity-trained checkpoints, we further study two complementary quantization-aware training (QAT) recipes: continued-QAT largely recovers performance at 4-bit weight quantization, while joint-QAT performs better for the more aggressive sub-4-bit target. To compare activation sparsity with dense low-bit quantization, we translate the sparsity-induced reduction in weight traffic into an equivalent bit-width. Under this unified weight-traffic metric, our 2.5- and 1.5-bit-equivalent models establish a new accuracy--traffic Pareto frontier: the 2.5-bit-equivalent models match dense 3-bit QAT accuracy, while the 1.5-bit-equivalent models approach dense 2-bit models and outperform dense ternary models by 1.3 pp on average. These results establish native-gate training as a practical approach for combining high FFN sparsity with low-bit weights.}

\date{\today}
\correspondence{\email{ruokaiyin@meta.com}}

\begin{document}

\maketitle

\section{Introduction}
\label{section:introduction}

Local inference of large language models (LLMs) is becoming increasingly important as on-device personal intelligence receives growing attention~\citep{liu2024mobilellm,agarwal2025gpt,chen2026mobilemoe}. Local LLM serving is typically bounded by single-user autoregressive decoding, whose speed is limited by moving model weights rather than by peak arithmetic throughput. Feed-forward networks (FFNs) account for a large fraction of these weights, even though only a token-dependent subset of their intermediate channels contributes meaningfully to each output token~\citep{liu2023deja}.

Activation sparsity exploits this property by loading only the weights for the active input/output channels~\citep{liu2024training,szatkowski2026universal}. The key challenge is to identify these channels early enough to reduce weight traffic without degrading the quality of the dense model.

Selecting these channels is particularly challenging in SwiGLU-based FFNs. The post-gating activation, i.e., the input to the down projection, provides the most accurate channel-importance signal~\citep{szatkowski2026universal}.
However, it becomes available only after both the up and gate projections have been computed, saving weight traffic only in the down projection. 
An auxiliary predictor~\citep{liu2023deja} can select channels before evaluating any of the three FFN projections, but introduces additional parameters and inference-time overhead. Existing predictor-based approaches have also been demonstrated primarily on ReLU-based FFNs, where exact zeros provide a clear prediction target~\citep{shrestha2026polar, song2025prosparse}.
The native gate and up signals provide a natural middle ground: one branch is computed densely and used as the router to select a shared channel set for the other branch and the down projection.

Recent training-free studies expose an important limitation of gate-based routing~\citep{liu2024training,szatkowski2026universal}. In a SwiGLU FFN, the gate signal ignores the up-projection output; a channel with a small gate value can therefore remain important when its up activation has a large magnitude.
Existing methods address this problem in different ways. CATS applies training-free magnitude thresholding to the gate signal~\citep{leecats}, but its performance degrades substantially at high sparsity. Other methods replace SiLU with ReLU and retrain the model to induce intrinsic zeros~\citep{mirzadeh2024relu,song2025prosparse}, at the cost of modifying the pretrained model architecture.

While we observe the same limitation in the training-free setting—where the gate signal performs substantially worse than the up signal at high sparsity—we find that this ordering reverses under sparsity-aware training. Our method, \name, trains all model parameters end-to-end using the native gate as a fixed top-$k$ router. \name preserves the pretrained SwiGLU activation function and introduces no additional router parameters. Across models ranging from 1B to 8B parameters, \name achieves 75--85\% FFN sparsity with approximately 30B training tokens while remaining within 0.2--2.7 pp of the dense models.

We further find that, unlike observations in prior training-free settings~\citep{liu2024training,xu2025resting, yin2026duogpt}, post-training quantization (PTQ) does not preserve the performance of sparsity-trained checkpoints. We therefore study two complementary strategies: continued-QAT, which applies quantization-aware training (QAT) after sparsity-aware training, and joint-QAT, which trains activation sparsity and weight quantization together.

Finally, we convert activation sparsity into an equivalent weight-traffic bit-width, allowing us to compare sparse W$_4$ and mixed W$_{2\&4}$ \name models directly with dense sub-4-bit QAT models on a unified accuracy--traffic Pareto frontier. Our key contributions are as follows:

\begin{itemize}
    \item To the best of our knowledge, we present the first systematic study that converts pretrained SwiGLU LLMs into fixed-budget activation-sparse models by training the native gate as the router. \name updates the full model through sparsity-aware training while preserving the pretrained SwiGLU activation and introducing no auxiliary router parameters. With approximately 30B training tokens, \name achieves uniform per-token 75--85\% FFN channel sparsity across LLaMA-3 models ranging from 1B to 8B parameters, within 0.2--2.7 pp of the corresponding dense models.
    \item We show that the optimal native routing signal changes with training. Although the up signal performs substantially better on frozen models, the native gate becomes the strongest deployable signal after sparsity-aware training.
    \item We show that PTQ consistently degrades sparsity-trained checkpoints and develop two QAT recipes, continued-QAT and joint-QAT, to combine activation sparsity with low-bit weights. Evaluated under an equivalent weight-traffic bit-width, our 2.5- and 1.5-bit-equivalent \name models establish new Pareto-efficient operating points, achieving accuracy comparable to dense 3- and 2-bit QAT models while outperforming dense ternary models by 1.3 pp on average.
\end{itemize}

\section{Related Work}
\label{sec:related-work}

Activation sparsity exploits token-dependent activation patterns to reduce computation and weight traffic during memory-bound LLM decoding~\citep{liu2023deja,mirzadeh2024relu,liu2024training}. Existing methods differ mainly in where they obtain the sparsity signal and whether they modify the pretrained FFN architecture.

\begin{figure}[t]
    \centering
    \includegraphics[width=\linewidth]{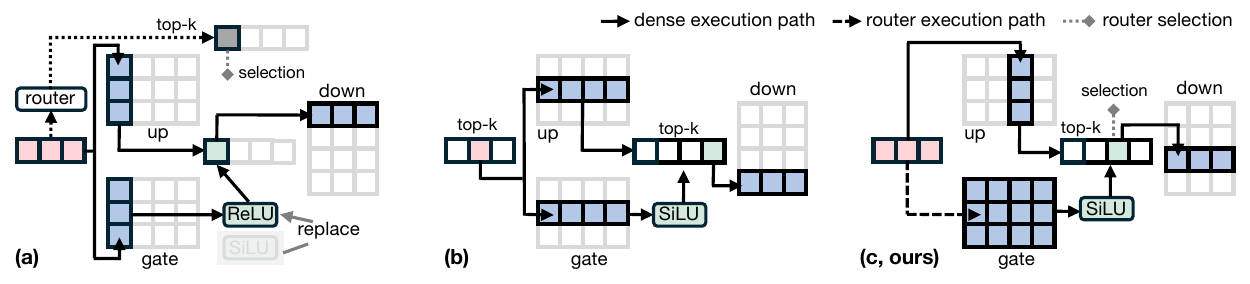}
    \caption{Comparison of different FFN activation-sparsity strategies: (a) auxiliary prediction with activation modification, (b) input activation sparsity, and (c) native-gate routing used by \name.}
    \label{fig:method-overview}
\end{figure}

\textbf{Auxiliary predictors and activation modification.}
ReLU-based Transformers naturally contain exact zeros in their FFN activations~\citep{li2022lazy}. DejaVu~\citep{liu2023deja} trains auxiliary predictors with a local binary classification objective to identify active channels before executing the FFN projections. However, auxiliary predictors introduce additional parameters and inference overhead, and extending them to SwiGLU FFNs remains challenging~\citep{shrestha2026polar}.
ReLUfication~\citep{mirzadeh2024relu} instead replaces SiLU with ReLU and continues training the model to produce intrinsic activation sparsity. ProSparse~\citep{song2025prosparse} and TwELL~\citep{cetin2026sparser} further increase this sparsity through additional regularization. These approaches can support high activation sparsity, but they change the activation function of the pretrained model. Figure~\ref{fig:method-overview}(a) illustrates this family of methods.

\textbf{Input activation sparsity.}
TEAL~\citep{liu2024training} sparsifies the inputs to the up, gate, and down projections using activation magnitude. R-Sparse~\citep{zhangr} and SPON~\citep{xu2025resting} compensate for the resulting information loss using low-rank residual paths and learned biases, respectively. DuoGPT~\citep{yin2026duogpt} combines input activation sparsity with per-channel weight sparsity. These methods operate without additional training. Q-Sparse~\citep{wang2024q} instead incorporates input activation sparsity during low-bit model training.
As illustrated in Figure~\ref{fig:method-overview}(b), applying input sparsity to every FFN projection sparsifies both the input and output dimensions of the up and gate projections. We refer to this as \textit{sparsity overlap}. Our experiments show that input-sparsity methods experience large accuracy degradation at high sparsity.

\textbf{Native-gate routing.}
CATS~\citep{leecats} is the closest prior work to ours. It uses the magnitude of the native post-SiLU gate activation and a calibrated threshold to select FFN channels without training. The gate projection is evaluated densely, and its output is reused to execute selected channels of the up and down projections. This approach preserves the original SwiGLU activation and requires no auxiliary router parameters.
However, prior studies find that the gate is less accurate than the up activation as a routing signal in the training-free setting~\citep{liu2024training,szatkowski2026universal}. \name studies a different setting: we apply fixed-budget top-$k$ routing and train all model parameters via sparsity-aware training.
We show that this training can reverse the signal ordering and turn the native gate into an effective router at high sparsity, without changing the pretrained SwiGLU activation.

\textbf{Activation sparsity and quantization.}
Several training-free sparsity methods report that activation sparsity can be combined with PTQ with little additional degradation~\citep{zhangr,liu2024training,xu2025resting,yin2026duogpt}. Q-Sparse~\citep{wang2024q} jointly trains input activation sparsity with binary or ternary weights, but does not study native-gate routing or the effect of quantizing a sparsity-trained checkpoint. ParetoQ~\citep{liu2026paretoq} provides strong dense low-bit QAT models across several precision levels. Our work studies how PTQ and QAT interact with sparsity-aware native-gate training and compares the resulting models with dense models trained with ParetoQ using a common equivalent weight-traffic metric.

\section{Preliminaries}
\label{sec:problem-setting}

\textbf{Notation. }
  Throughout this paper, column vectors are denoted by bold lowercase letters and matrices by bold uppercase letters. We denote a token hidden state by $\bn{x}\in\mathbb{R}^{d}$. The projection matrices are $\bn{W}^{\mathrm{up}},\bn{W}^{\mathrm{gate}}\in\mathbb{R}^{n\times d}$ and $\bn{W}^{\mathrm{down}}\in\mathbb{R}^{d\times n}$, where $d$ and $n$ denote the hidden and FFN intermediate dimensions, respectively.

\textbf{Router-based activation sparsity in SwiGLU FFNs. }
 A dense SwiGLU FFN computes
  \begin{equation}
      \bn{y}
      =
      \bn{W}^{\mathrm{down}}
      \left(
      \operatorname{SiLU}(\bn{W}^{\mathrm{gate}}\bn{x})
      \odot
      \bn{W}^{\mathrm{up}}\bn{x}
      \right).
      \label{eq:dense-swiglu}
  \end{equation}
We denote the gate activation, up activation, and post-gating activation by
  $\bn{g}=\operatorname{SiLU}(\bn{W}^{\mathrm{gate}}\bn{x})$,
  $\bn{u}=\bn{W}^{\mathrm{up}}\bn{x}$, and
  $\bn{h}=\bn{g}\odot\bn{u}$, respectively.
We apply activation sparsity by multiplying the post-gating activation $\bn{h}$ element-wise by a binary mask $\bn{m}\in\{0,1\}^n$. The post-gating signal $\bn{h}$ contains information from both branches, but it becomes available too late to skip either input projection. We therefore consider the native gate and up activations as early routing signals. For gate-based routing, we select
\begin{equation}
    {\mathcal{I}}_g=\operatorname{top\text{-}k}(|\bn{g}|,k)
    \label{eq:gate-select-problem}
\end{equation}
where $k$ is the number of retained intermediate channels. The resulting sparse FFN computation is:
\begin{equation}
    \bn{y}=\bn{W}^{\mathrm{down}}[:,\mathcal{I}_g]
    \left(\bn{g}[\mathcal{I}_g]\odot
    \bn{W}^{\mathrm{up}}[\mathcal{I}_g,:]\bn{x}\right).
    \label{eq:gate-routed-execution}
\end{equation}
The gate projection is evaluated densely to obtain $\mathcal{I}_g$, while only the selected gate values and corresponding up and down weights participate in the remaining computation. Up-based routing is defined analogously by exchanging the roles of the gate and up signals. We define the FFN channel sparsity as $s=1-k/n$.

\section{\name}
\label{sec:gate-routing}

\subsection{Gate vs. Up Routing without Training}
Prior work finds that the gate activation is less robust to sparsification than the up activation in frozen SwiGLU models~\citep{liu2024training,szatkowski2026universal}. Because $\bn{h}_i=\bn{g}_i\bn{u}_i$, ranking channels using $|\bn{g}|$ alone discards channels whose gate activations are small but whose up activations have large magnitudes. Consequently, gate-based routing introduces a larger FFN output error than up-based routing.

\begin{figure}[h]
\begin{center}
\def\arraystretch{0.5}
\begin{tabular}{@{}c@{}c}
\includegraphics[width=0.42\linewidth]{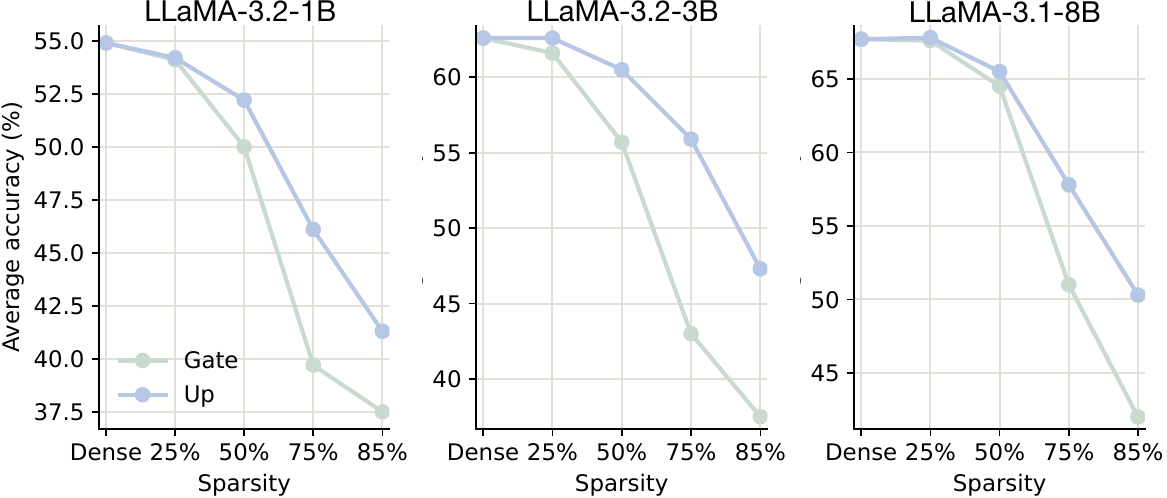} &
\includegraphics[width=0.55\linewidth]{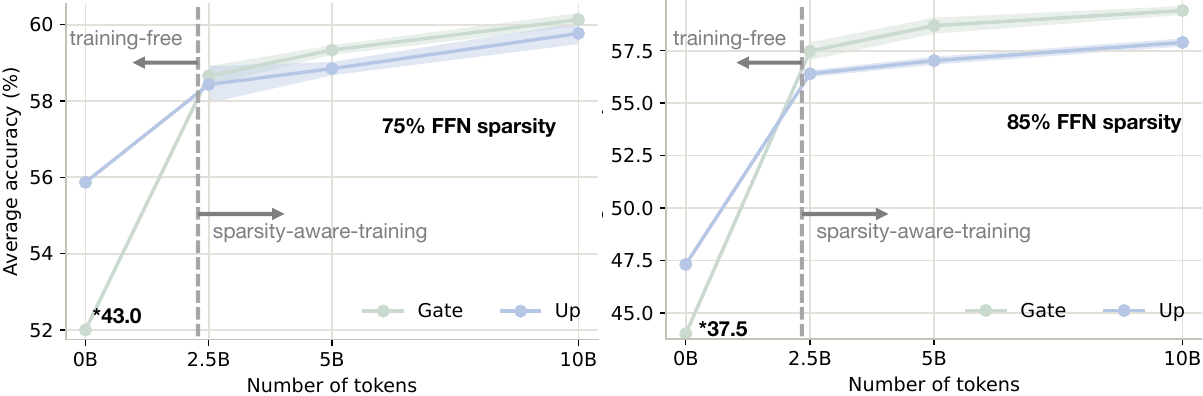} 
\\
{ (a)} & { (b)} 
\end{tabular}
\caption{(a) Training-free average zero-shot accuracy using gate- and up-based routing across FFN sparsity levels. (b) Average zero-shot accuracy during sparsity-aware training with gate- and up-based routing at 75\% and 85\% sparsity on LLaMA-3.2-3B, averaged over 4 seeds. For illustration purposes, we clip the 0B gate points.}
\label{fig:gate_vs_up}
\end{center}
\end{figure}

We verify this behavior across model sizes and sparsity levels. Figure~\ref{fig:gate_vs_up}(a) shows that the up signal begins to substantially outperform the gate signal at 50\% sparsity, and the gap widens at higher sparsity. On LLaMA-3.2-3B at 85\% sparsity, up-based routing outperforms gate-based routing by 9.8 pp in average accuracy. These results confirm that the native gate is an ineffective router when the pretrained model is kept frozen. We next examine whether this ordering persists once the model adapts to sparse execution.

\subsection{Activation-Sparsity-Aware Training}
\label{sec:asat}

Rather than requiring the routing signal to approximate the frozen FFN activation pattern, we train the full model end-to-end with routed FFNs enabled. Starting from pretrained parameters $\theta_0$, we select a target sparsity $s$ and minimize the causal language-modeling objective:
\begin{equation}
    \mathcal{L}(\theta;s)
    =
    \mathbb{E}_{z_{1:T}\sim\mathcal{D}}
      \left[-\frac{1}{T}\sum_{t=1}^{T}\log p_{\theta,s}(z_t\mid z_{<t})\right],
    \label{eq:sparse-cpt-objective}
\end{equation}
where $z_{1:T}$ is a sequence of $T$ tokens sampled from the training
distribution $\mathcal{D}$, and
$p_{\theta,s}$ denotes the next-token distribution under sparsity $s$. 

During training, we compute the gate and up projections densely and construct the gate-based index set $\mathcal{I}_g$ using Equation~\ref{eq:gate-select-problem}. We then form the binary mask $\bn m$ from $\mathcal{I}_g$ and apply it to the post-gating activation:
\begin{equation}
\begin{aligned}
    \bn{y} &= \bn{W}^{\mathrm{down}}(\bn{h}\odot\bn{m}), \text{where } 
    m_i =
    \begin{cases}
        1, & i\in\mathcal{I}_g\\
        0, & i\notin\mathcal{I}_g
    \end{cases}.
\end{aligned}
    \label{eq:train-fwd-swiglu}
\end{equation}
Equation~\ref{eq:train-fwd-swiglu} produces the exact same output as Equation~\ref{eq:gate-routed-execution}, while allowing training to leverage standard dense matrix multiplications without explicitly gathering rows and columns.

For backpropagation, let $\overline{\bn y}=\partial\mathcal{L}/\partial\bn y$ and $\bn q=(\bn W^{\mathrm{down}})^\top\overline{\bn y}$. For a mask fixed during the forward pass, the activation gradients are $\partial\mathcal{L}/\partial\bn g=\bn u\odot\bn m\odot\bn q$ and $\partial\mathcal{L}/\partial\bn u=\bn g\odot\bn m\odot\bn q$.
The down-projection gradient is $\partial\mathcal{L}/\partial\bn W^{\mathrm{down}}=\overline{\bn y}(\bn h\odot\bn m)^\top$. Thus, selected channels receive the standard gradient of the masked network, whereas dropped channels receive zero gradient for that token, but remain eligible to be selected and updated by subsequent tokens. We stop gradients through the hard top-$k$ operation and require neither a pseudo-gradient~\citep{hou2025instruction} nor an auxiliary routing loss. We discuss alternative pseudo-gradient formulations in Appendix~\ref{appendix:topk}.

\subsection{Gate vs. Up Routing after Training}

We next compare gate- and up-based routing under matched sparsity-aware training. Figure~\ref{fig:gate_vs_up}(b) reports the mean across 4 seeds (with shaded min--max ranges) on LLaMA-3.2-3B and shows that training rapidly closes the large initial gap between the two signals.

At 75\% sparsity, training completely eliminates the 12.9 pp initial deficit of gate-based routing, bringing both signals to parity with overlapping seed ranges. At 85\% sparsity, the ordering decisively reverses: gate-based routing overtakes up-based routing after approximately 2.5B training tokens and widens its lead to 1.5 pp in average accuracy at 10B tokens. Thus, the advantage of gate-based routing emerges most clearly in the high-sparsity regime where its frozen-model deficit was largest. These results demonstrate that routing quality in a frozen model is not indicative of routing capacity after sparsity-aware training. We therefore adopt the native gate as the router for all subsequent \name experiments.

\section{Activation Sparsity with Quantization}
\label{sec:sparsity-quantization}

\subsection{PTQ after Sparsity-Aware Training}
\label{sec:quant-coupling}

Prior training-free studies report that activation sparsity and post-training weight quantization can be combined with minimal additional degradation~\citep{zhangr,liu2024training,yin2026duogpt,xu2025resting}. However, this property does not transfer once the model parameters have adapted to sparse execution. As shown in Section~\ref{sec:experiments}, applying W$_4$ GPTQ~\citep{frantar2022gptq} directly to our sparsity-trained checkpoints reduces average accuracy by 0.7--1.3 pp, and extending BF16 sparsity-aware training by an additional 5B tokens prior to GPTQ does not mitigate this loss.

Because weight quantization simultaneously perturbs both the top-$k$ routing decisions and the retained projection values, post-hoc calibration is insufficient to preserve the learned sparse representations. We therefore incorporate quantization directly into sparsity-aware training.

\subsection{Continued-QAT and Joint-QAT}
\label{sec:quant-strategies}

We consider two weight-only quantization targets: uniform 4-bit quantization (W$_4$), and a more aggressive mixed-precision configuration (W$_{2\&4}$) in which the sparsely fetched up and down projections remain at 4 bits while all densely fetched quantized linear layers—including the gate projection—use 2 bits.

We develop two complementary QAT recipes tailored to these targets. \textit{Continued-QAT} initializes from a BF16 \name checkpoint, enables fake weight quantization while retaining gate-based top-$k$ routing, and continues training for a short budget (e.g., 5B tokens). This recipe allocates the majority of the training budget to BF16 sparsity-aware training and uses a brief QAT stage to adapt the router and weights to low precision. We find that continued-QAT is highly effective for the W$_4$ target.

\textit{Joint-QAT} initializes from the pretrained dense checkpoint and enables both gate-based activation sparsity and fake weight quantization from the first training step, allowing the model to co-adapt its routing patterns and quantized weights throughout training. Under a matched total token budget, joint-QAT outperforms continued-QAT on the more aggressive W$_{2\&4}$ target. Together, these two recipes provide flexible strategies depending on the target bit-width and available QAT budget.

\subsection{Equivalent Bit-Width and Weight Traffic}
\label{sec:router-aware-precision}

Activation sparsity and weight quantization reduce memory traffic along orthogonal axes: quantization reduces the bits transferred per fetched weight, whereas activation sparsity reduces the fraction of up and down weights fetched per token. To evaluate both approaches under a common weight-traffic metric, we convert sparsity and quantization into an equivalent bit-width.

Let $p_q$ denote the fraction of model parameters that are quantized and fetched densely, $p_s$ the fraction that are both quantized and sparsely fetched, and $p_o$ the fraction that remains dense in BF16, such that $p_q+p_s+p_o=1$. Let $b_q$ and $b_s$ denote the bit-widths of the first two groups. For an \name model at FFN sparsity $s$, we define its equivalent bit-width $b_{\mathrm{eq}}(s)$ as the uniform bit-width of the quantized parameter fraction ($1-p_o$) in a dense reference model that incurs the same per-token weight traffic. Because the $p_o$ parameters remain in BF16 in both models, equating their weight-matrix traffic gives
\begin{equation}
      16p_o+(1-p_o)b_{\mathrm{eq}}(s)
      =
      16p_o+p_qb_q+p_sb_s(1-s),
  \end{equation}
  and therefore
  \begin{equation}
      b_{\mathrm{eq}}(s)
      =
      \frac{p_qb_q+p_sb_s(1-s)}{1-p_o}.
      \label{eq:model-equivalent-bits}
  \end{equation}
Here, the dense gate and attention projections belong to $p_q$, while the sparsely fetched up and down projections belong to $p_s$. For LLaMA-3.2-3B, combining 75\% FFN sparsity with W$_4$ and W$_{2\&4}$ quantization yields $b_{\mathrm{eq}}=2.5$ and
$b_{\mathrm{eq}}=1.5$, respectively. We denote these two configurations as \textbf{SGA-Sparse-2.5} and \textbf{SGA-Sparse-1.5} throughout our evaluation (using the 3B architecture as the nominal reference).

\section{Experiments}
\label{sec:experiments}

\subsection{Experimental Setup } We evaluate \name on LLaMA-3.2-1B, LLaMA-3.2-3B, and LLaMA-3.1-8B~\citep{grattafiori2024llama}, all of which use SwiGLU FFNs. We report perplexity on WikiText-2~\citep{merity2016pointer} with a sequence length of 2048 and average accuracy over eight zero-shot commonsense reasoning tasks: ARC-Easy, ARC-Challenge~\citep{clark2018think}, BoolQ~\citep{clark2019boolq}, HellaSwag~\citep{zellers2019hellaswag}, OpenBookQA~\citep{mihaylov2018can}, PIQA~\citep{bisk2020piqa}, Social IQa~\citep{sap2019social}, and WinoGrande~\citep{sakaguchi2021winogrande}. We additionally evaluate generalization to Qwen-3.5-4B~\citep{qwen3.5} and report five-shot MMLU~\citep{hendrycks2020measuring} and GSM8K~\citep{cobbe2021training} results in the ablation study.

We initialize each model from its pretrained checkpoint and apply gate-based routing to every FFN block. We train on FineWeb-100BT~\citep{penedo2024fineweb} for up to 60,000 steps, corresponding to approximately 30B tokens. We use AdamW~\citep{adamw} with a weight decay of 0.05 and train on 32 GPUs with a batch size of 8 per GPU. Additional hyperparameters are provided in Table~\ref{tab:training-hparams}. For W$_4$ and W$_3$ QAT, we use learned quantization scales~\citep{lsq} with a group size of 128. For W$_2$ and ternary quantization, we use the SEQ formulation from ParetoQ~\citep{liu2026paretoq}, also with a group size of 128. Embedding and output-layer weights remain in BF16. Additional QAT settings are provided in Table~\ref{tab:training-hparams-qat}.

We compare against controlled top-$k$ implementations of the training-free methods CATS~\citep{leecats}, TEAL~\citep{liu2024training}, and R-Sparse~\citep{zhangr}, and the training-based methods Q-Sparse~\citep{wang2024q}, ReLUfication~\citep{mirzadeh2024relu}, and TwELL~\citep{cetin2026sparser}. Except for TwELL, we implement and evaluate all baselines in the same framework. To ensure deterministic weight traffic and an apples-to-apples comparison at fixed sparsity, we evaluate all baselines using exact per-token top-$k$ selection rather than variable-sparsity thresholding.

\begin{table}[h]
\caption{Comparison of 75\% and 85\% FFN sparsity on LLaMA-3.2-1B, 3.2-3B, and 3.1-8B models, against training-free baselines.}
\begin{adjustbox}{max width=\linewidth}
    \begin{tabular}{l||l|c|ccccccccc}
    
    \toprule\toprule

    \textbf{Model} & \textbf{Method} & \textbf{Wiki2$(\downarrow)$}& \textbf{ARC-C} & \textbf{ARC-E} & \textbf{BoolQ} & \textbf{HS} & \textbf{OBQA} & \textbf{PIQA} & \textbf{SIQA} & \textbf{WG} & \textbf{Avg Acc.$(\uparrow)$} \\
    \midrule
    \multirow{9}{*}{LLaMA-3.2-1B}& \cellcolor{lightgray}Dense & \cellcolor{lightgray}9.7 & \cellcolor{lightgray}36.4 & \cellcolor{lightgray}60.4 & \cellcolor{lightgray}63.9 & \cellcolor{lightgray}63.6 & \cellcolor{lightgray}36.8 & \cellcolor{lightgray}74.5 & \cellcolor{lightgray}42.9 &\cellcolor{lightgray}60.3 & \cellcolor{lightgray}54.9 \\
    
     &CATS(75\%)& 174.7 & 25.5 & 35.7 & 50.8 & 33.9 & 28.2 & 57.8 & 34.4 & 51.5 & 39.7\\
     &TEAL(75\%)& 30.7 & 26.2 & 41.6 & 52.9 & 41.8 & 28.8 & 62.0 & 37.8 & 51.1 & 42.8\\
    &R-Sparse(75\%)& 31.3 & 26.0 & 42.1 & \textbf{62.1} & 40.8 & 28.6 & 59.7 & 37.8 & 51.7 & 43.6\\
    &\cellcolor{sakura!70}\textbf{SGA-S}$_{30\text{BT}}$(75\%)& \cellcolor{sakura!70}\textbf{12.4} & \cellcolor{sakura!70}\textbf{38.1} & \cellcolor{sakura!70}\textbf{67.4} & \cellcolor{sakura!70}61.3 & \cellcolor{sakura!70}\textbf{59.0} & \cellcolor{sakura!70}\textbf{39.8} & \cellcolor{sakura!70}\textbf{73.2} & \cellcolor{sakura!70}\textbf{42.4} & \cellcolor{sakura!70}\textbf{56.6} & \cellcolor{sakura!70}\textbf{54.7}\\
    
    &CATS(85\%)& 17209.7 & 25.9 & 28.3 & 52.8 & 28.3 & 28.8 & 52.6 & 33.5 & 49.6 & 37.5\\
    &TEAL(85\%)& 487.1 & 24.9 & 30.5 & 50.7 & 28.9 & 27.4 & 54.1 & 35.4 & 50.6 & 37.8\\
    &R-Sparse(85\%)& 328.3 & 24.2 & 31.3 & 59.3 & 29.3 & 23.6 & 53.7 & 34.4 & 49.9 & 38.2\\ 
    & \cellcolor{sakura!70}\textbf{SGA-S}$_{30\text{BT}}$(85\%) & \cellcolor{sakura!70}\textbf{12.6} & \cellcolor{sakura!70}\textbf{37.4} & \cellcolor{sakura!70}\textbf{67.5} & \cellcolor{sakura!70}\textbf{60.1} & \cellcolor{sakura!70}\textbf{58.4} & \cellcolor{sakura!70}\textbf{38.8} & \cellcolor{sakura!70}\textbf{72.7} & \cellcolor{sakura!70}\textbf{40.7} & \cellcolor{sakura!70}\textbf{57.5} & \cellcolor{sakura!70}\textbf{54.1} \\
    \midrule
    
     \multirow{9}{*}{LLaMA-3.2-3B}& \cellcolor{lightgray}Dense & \cellcolor{lightgray}7.8 & \cellcolor{lightgray}45.7 & \cellcolor{lightgray}71.6 & \cellcolor{lightgray}72.9 & \cellcolor{lightgray}73.6 & \cellcolor{lightgray}42.8 & \cellcolor{lightgray}77.5 & \cellcolor{lightgray}47.1 & \cellcolor{lightgray}69.5 & \cellcolor{lightgray}62.6 \\
    
     &CATS(75\%)& 128.8 & 30.0 & 40.0 & 47.8 & 45.6 & 31.6 & 60.9 & 35.9 & 52.1 & 43.0\\
     &TEAL(75\%)& 17.7 & 33.7 & 51.6 & 54.5 & 55.3 & 35.4 & 67.6 & 41.3 & 55.9 & 49.4\\
     &R-Sparse(75\%)& 18.1 & 33.0 & 51.3 & 63.9 & 54.3 & 30.6 & 68.6 & 40.6 & 55.6 & 49.7\\
     &\cellcolor{sakura!70}\textbf{SGA-S}$_{30\text{BT}}$(75\%)& \cellcolor{sakura!70}\textbf{9.6} & \cellcolor{sakura!70}\textbf{47.1} & \cellcolor{sakura!70}\textbf{73.7} & \cellcolor{sakura!70}\textbf{71.0} & \cellcolor{sakura!70}\textbf{68.7} & \cellcolor{sakura!70}\textbf{44.0} & \cellcolor{sakura!70}\textbf{77.5} & \cellcolor{sakura!70}\textbf{45.7} & \cellcolor{sakura!70}\textbf{64.5} & \cellcolor{sakura!70}\textbf{61.5}\\
     
    &CATS(85\%)& 5122.2 & 27.3 & 32.2 & 43.2 & 30.7 & 28.2 & 55.7 & 33.3 & 49.3 & 37.5\\
    &TEAL(85\%)& 167.1 & 22.8 & 34.8 & 52.0 & 33.1 & 26.6 & 56.1 & 35.9 & 50.5 & 39.0\\ 
    &R-Sparse(85\%)& 93.9 & 23.4 & 35.5 & 60.6 & 34.7 & 26.8 & 57.7 & 36.1 & 52.7 & 40.9\\
    & \cellcolor{sakura!70}\textbf{SGA-S}$_{30\text{BT}}$(85\%) & \cellcolor{sakura!70}\textbf{9.7} & \cellcolor{sakura!70}\textbf{46.3} & \cellcolor{sakura!70}\textbf{73.2} & \cellcolor{sakura!70}\textbf{70.9} & \cellcolor{sakura!70}\textbf{68.1} & \cellcolor{sakura!70}\textbf{43.2} & \cellcolor{sakura!70}\textbf{77.0} & \cellcolor{sakura!70}\textbf{44.9} & \cellcolor{sakura!70}\textbf{65.8} & \cellcolor{sakura!70}\textbf{61.2} \\
    \midrule
    
    \multirow{9}{*}{LLaMA-3.1-8B}&\cellcolor{lightgray}Dense & \cellcolor{lightgray}6.2 & \cellcolor{lightgray}53.3 & \cellcolor{lightgray}81.1 & \cellcolor{lightgray}82.1 & \cellcolor{lightgray}78.9 & \cellcolor{lightgray}44.4 & \cellcolor{lightgray}81.1 & \cellcolor{lightgray}47.0 & \cellcolor{lightgray}73.5 & \cellcolor{lightgray}67.7 \\
    &CATS(75\%)& 28.4 & 34.0 & 50.6 & 55.4 & 63.5 & 37.4 & 72.0 & 40.2 & 54.5 & 51.0\\
    &TEAL(75\%)& 12.5 & 37.8 & 59.5 & 69.6 & 65.5 & 40.0 & 74.5 & 42.9 & 60.3 & 56.3\\
     &R-Sparse(75\%)& 13.0 & 39.2 & 63.2 & 71.5 & 64.8 & 38.2 & 73.6 & 43.0 & 62.4 & 57.0\\
      &\cellcolor{sakura!70}\textbf{SGA-S}$_{30\text{BT}}$(75\%)& \cellcolor{sakura!70}\textbf{7.9} & \cellcolor{sakura!70}\textbf{52.6} & \cellcolor{sakura!70}\textbf{78.2} & \cellcolor{sakura!70}\textbf{77.6} & \cellcolor{sakura!70}\textbf{75.9} & \cellcolor{sakura!70}\textbf{46.4} & \cellcolor{sakura!70}\textbf{79.4} & \cellcolor{sakura!70}\textbf{47.2} & \cellcolor{sakura!70}\textbf{70.6} & \cellcolor{sakura!70}\textbf{66.0}\\
    &CATS(85\%)& 956.8 & 29.0 & 36.9 & 48.8 & 41.6 & 31.0 & 62.1 & 35.3 & 51.5 & 42.0\\
    &TEAL(85\%)& 46.6 & 27.2 & 41.1 & 52.1 & 42.1 & 29.6 & 59.6 & 36.7 & 55.3 & 43.0\\
    &R-Sparse(85\%)& 46.7 & 28.3 & 43.0 & 63.9 & 44.0 & 27.4 & 61.5 & 38.4 & 51.7 & 44.8\\
    & \cellcolor{sakura!70}\textbf{SGA-S}$_{30\text{BT}}$(85\%)& \cellcolor{sakura!70}\textbf{8.1} & \cellcolor{sakura!70}\textbf{49.9} & \cellcolor{sakura!70}\textbf{77.4} & \cellcolor{sakura!70}\textbf{76.6} & \cellcolor{sakura!70}\textbf{75.0} & \cellcolor{sakura!70}\textbf{45.0} & \cellcolor{sakura!70}\textbf{79.3} & \cellcolor{sakura!70}\textbf{46.9} & \cellcolor{sakura!70}\textbf{69.4} & \cellcolor{sakura!70}\textbf{65.0} \\

    \bottomrule\bottomrule
    \end{tabular}
  \end{adjustbox}
\label{tab_main_trainingfree}
\end{table}

\subsection{Main Results }
We first compare \name with the training-free baselines in Table~\ref{tab_main_trainingfree}. At 75\% and 85\% FFN sparsity, every training-free baseline loses more than 10 pp of average accuracy relative to the corresponding dense model. R-Sparse performs best among these baselines, aided by its additional low-rank residual paths. Nevertheless, at 85\% sparsity, it remains 21.7 and 22.9 points below the
dense 3B and 8B models, respectively. Sparsity-aware training substantially reduces these gaps. At 75\% sparsity, \name remains within 0.2, 1.1, and 1.7 points of the dense 1B, 3B, and 8B models, respectively. At 85\% sparsity, the corresponding gaps are 0.8, 1.4, and 2.7 points. The larger gaps at 8B suggest that larger models may require more training than the fixed 30B-token budget used here.

\begin{table}[h]
\centering \footnotesize
\begin{minipage}[h]{0.5\linewidth}
\centering
\caption{Comparison with activation-sparsity-aware training baselines at 85\% sparsity. All methods use a 20B-token training budget. *Evaluated directly on the provided 20BT-trained checkpoint. All other methods use LLaMA-3.2.}
\label{tab_main_training}
\begin{adjustbox}{max width=\linewidth}
\begin{tabular}{c||l|cc}
\toprule\toprule
\textbf{Model Size} & \textbf{Method} & \textbf{Wiki2$(\downarrow)$} & \textbf{Avg Acc.$(\uparrow)$} \\
\midrule
\multirow{4}{*}{1B}
& Q-Sparse$_{20\text{BT}}$(85\%) & 14.3 & 51.6 \\
& ReLUfication$_{20\text{BT}}$(85\%) & 18.8 & 51.3 \\
& TwELL$_{20\text{BT}}$(85\%)* & 15.4 & 50.3 \\
& \cellcolor{sakura!70}\textbf{SGA-S}$_{20\text{BT}}$(85\%) & \cellcolor{sakura!70}\textbf{12.8} & \cellcolor{sakura!70}\textbf{53.9} \\
\midrule\midrule
\multirow{3}{*}{3B}
& Q-Sparse$_{20\text{BT}}$(85\%) & 10.9 & 57.6 \\
& ReLUfication$_{20\text{BT}}$(85\%) & 10.2 & 59.8 \\
& \cellcolor{sakura!70}\textbf{SGA-S}$_{20\text{BT}}$(85\%) & \cellcolor{sakura!70}\textbf{9.9} & \cellcolor{sakura!70}\textbf{60.8} \\
\bottomrule\bottomrule
\end{tabular}
\end{adjustbox}
\end{minipage}\hfill
\begin{minipage}[h]{0.46\linewidth}
\centering
\caption{W$_4$ quantization on 75\%-sparse \name models. We compare GPTQ after 30B or 35B tokens of sparse training with continued-QAT using 30B tokens of sparse training followed by 5B QAT.}
\label{tab_main_w4quant}
\begin{adjustbox}{max width=\linewidth}
\begin{tabular}{l|l||cc}
\toprule\toprule
\textbf{Model} & \textbf{Method} & \textbf{Wiki2$(\downarrow)$} & \textbf{Avg Acc.$(\uparrow)$}\\
\midrule
\multirow{3}{*}{3.2-1B}
&\textbf{SGA-S}$_{30\text{BT}}$(75\%) + GPTQ(W$_{4}$) & 13.1 & 53.8\\
&\textbf{SGA-S}$_{35\text{BT}}$(75\%) + GPTQ(W$_{4}$) & 13.0 & 54.1\\
&\cellcolor{sakura!70}\textbf{SGA-S}$_{30\text{BT}}$(75\%) + QAT$_{5\text{BT}}$(W$_{4}$) & \cellcolor{sakura!70}\textbf{12.6} & \cellcolor{sakura!70}\textbf{54.9}\\
\midrule
\multirow{3}{*}{3.2-3B}
&\textbf{SGA-S}$_{30\text{BT}}$(75\%) + GPTQ(W$_{4}$) & 10.8 & 60.2\\
&\textbf{SGA-S}$_{35\text{BT}}$(75\%) + GPTQ(W$_{4}$) & 10.0 & 59.9\\
&\cellcolor{sakura!70}\textbf{SGA-S}$_{30\text{BT}}$(75\%) + QAT$_{5\text{BT}}$(W$_{4}$) & \cellcolor{sakura!70}\textbf{9.8} & \cellcolor{sakura!70}\textbf{61.6}\\
\midrule
\multirow{3}{*}{3.1-8B}
&\textbf{SGA-S}$_{30\text{BT}}$(75\%) + GPTQ(W$_{4}$) & 8.2 & 65.3\\
&\textbf{SGA-S}$_{35\text{BT}}$(75\%) + GPTQ(W$_{4}$) & 8.2 & 65.2\\
&\cellcolor{sakura!70}\textbf{SGA-S}$_{30\text{BT}}$(75\%) + QAT$_{5\text{BT}}$(W$_{4}$) & \cellcolor{sakura!70}\textbf{8.0} & \cellcolor{sakura!70}\textbf{65.5}\\
\bottomrule\bottomrule
\end{tabular}
\end{adjustbox}
\end{minipage}
\end{table}

We next compare against training-based methods under a matched 20B-token budget in Table~\ref{tab_main_training}. \name achieves the best perplexity and average accuracy across both the 1B and 3B models.
Compared with Q-Sparse, it improves the 3B accuracy by
3.2 points at 85\% sparsity. We attribute part of this difference to the sparsity overlap introduced by applying input sparsity to consecutive FFN projections. At 85\% sparsity, \name outperforms ReLUfication and TwELL by at least 2.6 pp on the 1B model. On the 3B model, it outperforms ReLUfication by 1.0 pp while preserving the original SwiGLU activation.

\begin{wraptable}{r}{0.48\linewidth} 
\centering \footnotesize 
\captionof{table}{Comparison of continued-QAT and joint-QAT for 75\%-sparse mixed-W$_{2\&4}$ models under the same 30B-token budget.} \label{tab:comparison:qat_methods} 
\begin{adjustbox}{max width=\linewidth}
        \begin{tabular}{l|lcc}
    \toprule\toprule
         \textbf{Model} & \textbf{Method} & \textbf{Wiki2$(\downarrow)$} & \textbf{Avg Acc$(\uparrow)$}\\
        \midrule
        \multirow{2}{*}{3.2-1B}&Continued-QAT (25B + 5B) &14.8 &51.8\\
        &\cellcolor{sakura!70}Joint-QAT (30B) & \cellcolor{sakura!70}\textbf{13.4}&\cellcolor{sakura!70}\textbf{53.1}\\
        \midrule
        \multirow{2}{*}{3.2-3B}&Continued-QAT (25B + 5B)& 11.1 & 58.5\\
        &\cellcolor{sakura!70}Joint-QAT (30B)& \cellcolor{sakura!70}\textbf{10.3} &\cellcolor{sakura!70}\textbf{59.9}\\
        \midrule 
        \multirow{2}{*}{3.1-8B}&Continued-QAT (25B + 5B) &9.0 & 63.5\\
        &\cellcolor{sakura!70}Joint-QAT (30B)& \cellcolor{sakura!70}\textbf{8.6} &\cellcolor{sakura!70}\textbf{64.6}\\
    \bottomrule \bottomrule
    \end{tabular}
    \end{adjustbox}
\end{wraptable}

Table~\ref{tab_main_w4quant} compares W$_4$ GPTQ~\citep{frantar2022gptq} with continued-QAT for 75\%-sparse \name checkpoints. Applying GPTQ with a group size of 128 directly after 30B tokens of sparsity-aware training reduces average accuracy by 0.7--1.3 points. Performing another 5B tokens of BF16 sparsity-aware training before GPTQ does not consistently reduce this loss. In contrast, continued-QAT recovers the BF16 sparse accuracy on the 1B and 3B models and remains within 0.5 pp on the 8B model.

We next compare the two QAT recipes for the more aggressive
1.5-bit-equivalent target. Continued-QAT uses 25B tokens of BF16 sparsity-aware training followed by 5B tokens of QAT, whereas joint-QAT enables sparsity and quantization throughout all 30B training tokens. As shown in Table~\ref{tab:comparison:qat_methods}, joint-QAT improves average accuracy by 1.3, 1.4, and 1.1 points on the 1B, 3B, and 8B models, respectively. It also reduces perplexity at every model size. These results show that the aggressive mixed-W$_{2\&4}$ target benefits from learning the routing pattern and quantized weights jointly.

\begin{figure}[h]
    \centering
    \includegraphics[width=0.9\linewidth]{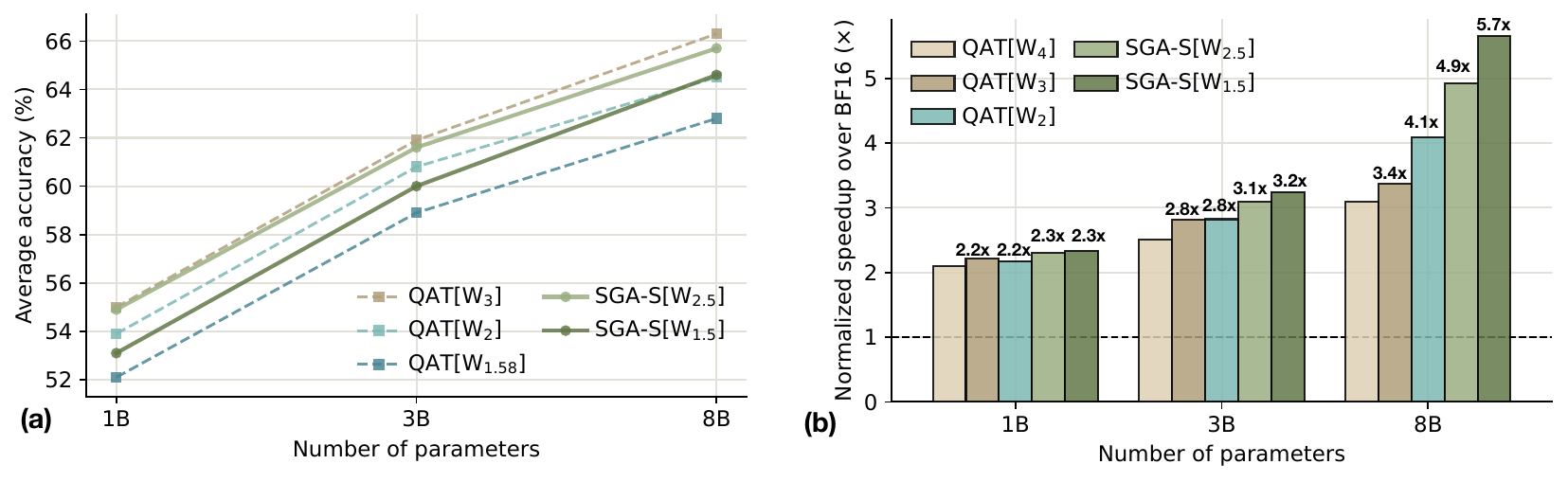}
    \caption{(a) Average zero-shot accuracy across model sizes between dense QAT models and quantized \name (SGA-S) models. (b) FFN GEMV speedup during single-token decoding over dense BF16 baseline on an NVIDIA A100 80GB GPU.}
    \label{fig:qat_comparison}
\end{figure}

Finally, we compare quantized \name models with dense low-bit QAT models using equivalent weight traffic. As shown in Figure~\ref{fig:qat_comparison}(a), the \textbf{SGA-Sparse-2.5} models achieve accuracy comparable to dense W$_3$ models across model sizes ranging from 1B to 8B parameters. The 1.5-bit-equivalent models similarly approach the accuracy of dense W$_2$ models and consistently outperform dense ternary models by 1.3 pp on average. These results introduce new Pareto-efficient operating points among the evaluated models. Figure~\ref{fig:qat_comparison}(b) further evaluates kernel-level GPU efficiency by benchmarking the summed GEMV kernel latency of the dense gate and sparse up and down projections. Both W$_4$+75\% sparsity and W$_{2\&4}$+75\% sparsity achieve higher projection-level speedups than the dense W$_2$ baseline.

\begin{figure}[h]
\begin{center}
\def\arraystretch{0.5}
\begin{tabular}{@{}c@{}c}
\includegraphics[width=0.49\linewidth]{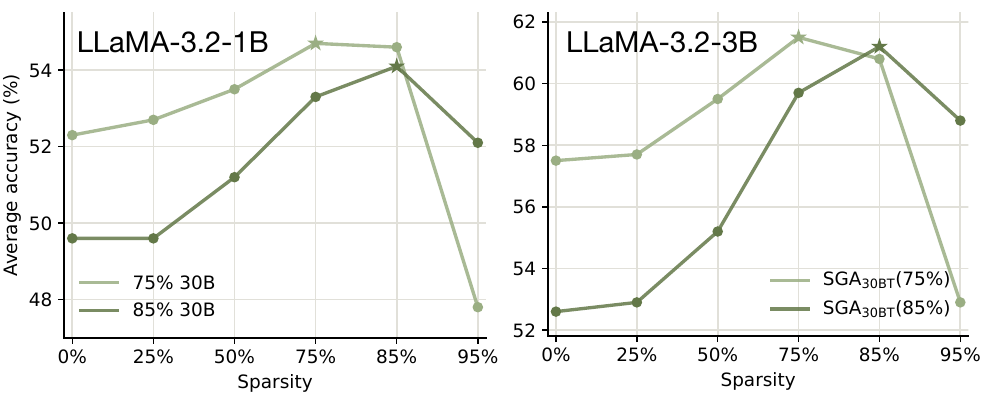} &
\includegraphics[width=0.49\linewidth]{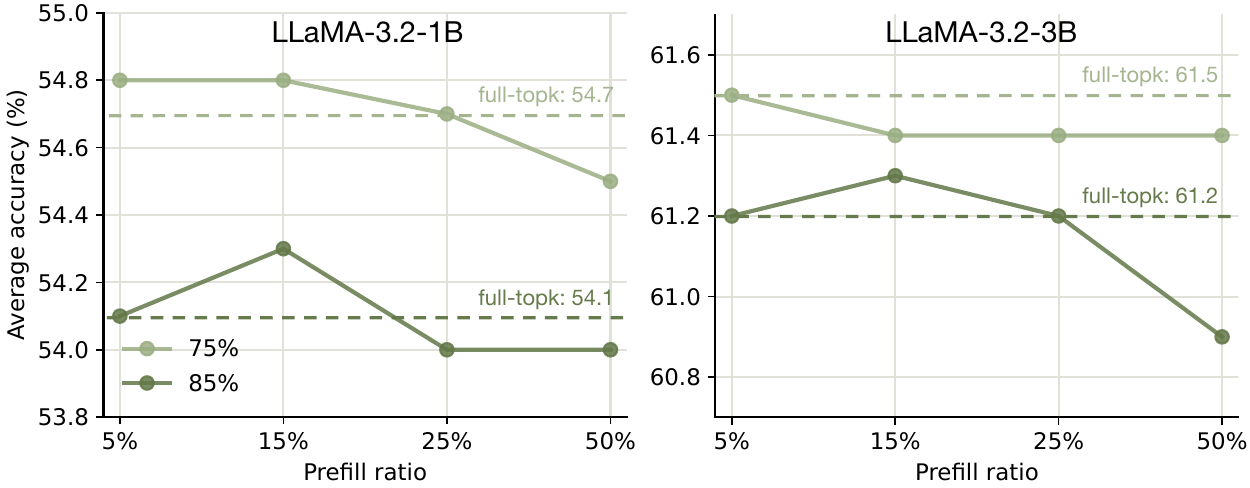} 
\\
{ (a)} & { (b)} 
\end{tabular}
\caption{(a) Evaluation of sparsity-trained checkpoints across varying test-time sparsity levels. (b) Accuracy using threshold-based routing during prefill across prefill ratios from 5\% to 50\%~\citep{liu2024training,zhangr}.}\vspace{-3mm}
\label{fig:ablation:sparsity_level}
\end{center}
\end{figure}

\subsection{Ablation Studies }
Figure~\ref{fig:ablation:sparsity_level}(a) evaluates each checkpoint at sparsity levels different from its training target. Each model performs best near the sparsity level used during training, and accuracy decreases as the evaluation sparsity moves away from this target. Nevertheless, evaluating the 85\%-trained checkpoints at 95\% sparsity reduces average accuracy by only approximately two points on the 1B and 3B models. This suggests that further training from an 85\%-sparse checkpoint may provide an efficient path toward more extreme sparsity.
Furthermore, according to Figure~\ref{fig:ablation:sparsity_level}(a), applying dense execution to prefill tokens also creates a mismatch with the sparse computation used during training. We therefore apply an offline calibrated threshold during prefill. As shown in Figure~\ref{fig:ablation:sparsity_level}(b), threshold-based prefill preserves accuracy across the evaluated settings without performing a top-$k$ operation for every prefill token. The resulting number of active channels can vary across tokens, but our primary deployment acceleration target is single-token decoding, where we retain fixed-budget top-$k$ routing.
Table~\ref{tab:ablation:qwen} extends our evaluation to Qwen-3.5-4B. With only 10B training tokens, gate-based \name remains within 1.0 and 1.9 pp of the dense model at 75\% and 85\% sparsity, respectively. It also outperforms up-based routing by 0.9 and 1.3 pp at the two sparsity levels. These results show that both \name and the training-induced preference for gate-based routing extend beyond the LLaMA family to another pretrained SwiGLU model family. Finally, Table~\ref{tab:ablation:reasoning} compares \name and Q-Sparse on LLaMA-3.2-3B models trained with 30B tokens on five-shot MMLU and GSM8K, where \name provides the largest gains at 85\% sparsity.

\begin{table}[h]
\centering \footnotesize
\begin{minipage}[h]{0.44\linewidth}
\centering
\caption{Comparison of gate- and up-based routing on Qwen-3.5-4B.}
\label{tab:ablation:qwen}
\begin{adjustbox}{max width=\linewidth}
\begin{tabular}{l||l|cc}
\toprule\toprule
\textbf{Model} & \textbf{Method} & \textbf{Wiki2$(\downarrow)$} & \textbf{Avg Acc.$(\uparrow)$} \\
\midrule
\multirow{5}{*}{Qwen-3.5-4B}
& \cellcolor{lightgray}Dense & \cellcolor{lightgray}8.0 & \cellcolor{lightgray}66.6 \\
& \textbf{SGA-S}$_{10\text{BT}}$(75\%)-up & 9.1 & 64.7 \\
& \cellcolor{sakura!70}\textbf{SGA-S}$_{10\text{BT}}$(75\%)-gate & \cellcolor{sakura!70}\textbf{8.8} & \cellcolor{sakura!70}\textbf{65.6} \\
& \textbf{SGA-S}$_{10\text{BT}}$(85\%)-up & 9.5 & 63.4 \\
& \cellcolor{sakura!70}\textbf{SGA-S}$_{10\text{BT}}$(85\%)-gate & \cellcolor{sakura!70}\textbf{8.9} & \cellcolor{sakura!70}\textbf{64.7} \\
\bottomrule\bottomrule
\end{tabular}
\end{adjustbox}
\end{minipage}\hfill
\begin{minipage}[h]{0.54\linewidth}
\centering
\caption{Performance comparison with Q-Sparse on few-shot reasoning tasks.}
\label{tab:ablation:reasoning}
\begin{adjustbox}{max width=\linewidth}
\begin{tabular}{l||ccc}
\toprule\toprule
\textbf{Method} & \textbf{5-shot MMLU} & \textbf{5-shot GSM8K} & \textbf{Avg. 0-shot tasks}\\
\midrule
Q-Sparse$_{30\text{BT}}$(75\%) & 39.9 & 9.0 & 59.4\\
\rowcolor{sakura!70} \textbf{SGA-S}$_{30\text{BT}}$(75\%) & \textbf{40.5} & \textbf{9.2} & \textbf{61.5}\\
Q-Sparse$_{30\text{BT}}$(85\%) & 33.2 & 5.8 & 58.5\\
\rowcolor{sakura!70} \textbf{SGA-S}$_{30\text{BT}}$(85\%) & \textbf{38.1} & \textbf{8.2} & \textbf{61.2}\\
\bottomrule\bottomrule
\end{tabular}
\end{adjustbox}
\end{minipage}
\end{table}

\section{Discussion and Future Work}
\label{sec:limitations}

We hypothesize that gate-based routing overtakes up-based routing during training because the smooth, bounded-below $\operatorname{SiLU}$ nonlinearity naturally adapts to channel gating, whereas routing via the linear up branch forces $\bn{W}^{\mathrm{up}}$ to simultaneously rank channels and carry unbounded value magnitudes. Similarly, our $\text{W}_{2\&4}$ mixed-precision design reflects this functional split: 2-bit weights suffice for the densely fetched gate projection to maintain relative top-$k$ ordering, whereas 4-bit weights preserve value fidelity on the sparsely fetched up and down channels. A deeper theoretical analysis of this routing reversal, alongside a broader exploration of mixed-precision bit allocations, remains an interesting direction for future study. Finally, our present study focuses on intermediate-channel sparsity in dense SwiGLU FFNs across GQA-based LLaMA-3 (1B to 8B) and GDN-based Qwen-3.5 (4B) models under a single pretraining distribution; extending our findings to ReLU-based FFNs, attention sparsity, and mixture-of-experts routing is a natural next step for future work.

\section{Conclusion}
\label{sec:conclusion}
We present \textbf{SGA-Sparse}, a method for converting pretrained SwiGLU LLMs into fixed-budget activation-sparse models without modifying the pretrained activation function or introducing auxiliary router parameters. By training the existing SwiGLU gate as a top-$k$ router through full-model sparsity-aware training, we show that the optimal routing signal changes with training: although the up activation is substantially stronger in frozen models, the native gate becomes the superior router after training. With approximately 30B training tokens, \name achieves uniform per-token 75--85\% FFN sparsity across LLaMA models from 1B to 8B parameters while remaining within 0.2--2.7 pp of their original dense checkpoints, and generalizes to other model families such as Qwen-3.5. We further demonstrate that weight quantization must be co-designed with sparsity-aware training and introduce two complementary QAT recipes: continued-QAT and joint-QAT. Evaluated under a unified equivalent weight-traffic bit-width, our 2.5-bit-equivalent models match dense W$_3$ accuracy, while our 1.5-bit-equivalent models approach dense W$_2$ performance and outperform dense ternary models by 1.3 pp on average. Together, these results establish native-gate training as a practical way to combine high activation sparsity with low-bit weights. More broadly, \name positions activation-sparsity-aware training alongside QAT as a promising model-optimization paradigm for efficient on-device LLM deployment.

\clearpage
\newpage
\bibliographystyle{assets/plainnat}
\bibliography{paper}

\clearpage
\newpage
\beginappendix

\section{Reproducibility Details}

Tables~\ref{tab:training-hparams} and~\ref{tab:training-hparams-qat} summarize the training configurations used for our main \name runs and the dense ParetoQ baselines, respectively. For the ParetoQ baselines, we use one BF16 quantization scale per group of 128 weights. Because group-wise scaling converges faster than the per-channel scaling used in the original work~\citep{liu2026paretoq}, we train the W$_2$ and ternary models for 20B tokens, at which point they match the accuracy reported in the original paper.

For continued-QAT, we use a peak learning rate of 1e-5 with all other hyperparameters remaining the same. For joint-QAT, we use the same learning rate as the corresponding BF16 sparsity-aware training run. For Qwen-3.5-4B, both gate- and up-based routing runs use a peak learning rate of 4e-5.
  
\begin{table*}[h]
    \centering
    \caption{Hyperparameters for the main sparsity-aware training runs. All models are trained on 32 GPUs with a sequence length of 2,048 and a linear learning-rate schedule.}
    \label{tab:training-hparams}
    \small
    \begin{adjustbox}{max width=\linewidth}
    \begin{tabular}{llccccccc}
        \toprule
        \toprule
        \textbf{Model} & \textbf{Method} & \textbf{Steps}
        & \textbf{Tokens/Step} & \textbf{Peak/Min. LR}
        & \textbf{WD} &\textbf{Warmup} & \textbf{Batch/GPU} & \textbf{Seed} \\
        \midrule
        \multirow{2}{*}{LLaMA-3.2-1B} & \textbf{SGA-S}$_{30\text{BT}}$(75\%) & 60000 & 524288 & 1e-4/0 & 5e-2 & 5\% & 8 & 428 \\
         & \textbf{SGA-S}$_{30\text{BT}}$(85\%) & 60000 & 524288 & 1e-4/0 & 5e-2 & 5\% & 8 & 428 \\
        \midrule
        \multirow{2}{*}{LLaMA-3.2-3B} & \textbf{SGA-S}$_{30\text{BT}}$(75\%) & 60000 & 524288 & 1e-4/0 & 5e-2 & 5\% & 8 & 428 \\
         & \textbf{SGA-S}$_{30\text{BT}}$(85\%) & 60000 & 524288 & 8e-5/0 & 5e-2 & 5\% & 8 & 428 \\
        \midrule
        \multirow{2}{*}{LLaMA-3.1-8B} & \textbf{SGA-S}$_{30\text{BT}}$(75\%) & 60000 & 524288 & 4e-5/0 & 5e-2 & 5\% & 8 & 428 \\
         & \textbf{SGA-S}$_{30\text{BT}}$(85\%) & 60000 & 524288 & 4e-5/0 & 5e-2 & 5\% & 8 & 428 \\
        \bottomrule
        \bottomrule
    \end{tabular}
    \end{adjustbox}
\end{table*}

\begin{table*}[h]
    \centering
    \caption{Hyperparameters for the dense ParetoQ QAT baselines. All models are trained on 32 GPUs with a sequence length of 2,048 and a linear learning-rate schedule. Following the original implementation, learnable quantization scales use zero weight decay and share the base learning rate of the weights.}
    
    \label{tab:training-hparams-qat}
    \small
    \begin{adjustbox}{max width=\linewidth}
    \begin{tabular}{llcccccccc}
        \toprule
        \toprule
        \textbf{Model} & \textbf{\# of Bits} & \textbf{Steps}
        & \textbf{Tokens/Step} & \textbf{Peak/Min. LR} & \textbf{Group Size}
        & \textbf{WD} &\textbf{Warmup} & \textbf{Batch/GPU} & \textbf{Seed} \\
        \midrule
        \multirow{3}{*}{1B/3B/8B} & W$_3$ & 20000 & 524288 & 2e-5/0 & 128 & 5e-2 & 5\% & 8 & 428 \\
        & W$_2$ & 40000 & 524288 & 4e-5/0 & 128 & 5e-2 & 5\% & 8 & 428 \\
        & W$_{1.58}$ & 40000 & 524288 & 4e-5/0 & 128 & 5e-2 & 5\% & 8 & 428 \\
        \bottomrule
        \bottomrule
    \end{tabular}
    \end{adjustbox}
\end{table*}

\section{Additional Method Details}

\subsection{Gradient Through Hard Top-$k$ Routing}
\label{appendix:topk}

For gate-based routing, let $\bn{r}=|\bn{g}|$ and $\bn{m}=M_k(\bn{r})$, where $[M_k(\bn{r})]_i=1$ if $i\in\operatorname{top\text{-}k}(\bn{r},k)$ and is zero otherwise. Let $\bn{q}=(\bn{W}^{\mathrm{down}})^\top\overline{\bn{y}}$. Since $\partial\mathcal{L}/\partial\bn{m}=\bn{h}\odot\bn{q}$, the gradient with respect to the gate activation can be decomposed as
\begin{equation}
\begin{aligned}
    \frac{\partial\mathcal{L}}{\partial\bn{g}}
    &=
    \underbrace{\bn{u}\odot\bn{m}\odot\bn{q}}_{
    \text{direct activation path}} \\
    &\quad+
    \underbrace{\operatorname{sgn}(\bn{g})\odot
    J_{M_k}(\bn{r})^\top(\bn{h}\odot\bn{q})}_{
    \text{routing path}},
\end{aligned}
    \label{eq:topk-routing-gradient}
\end{equation}
where $J_{M_k}=\partial\bn{m}/\partial\bn{r}$. The first term is the direct activation gradient used in Section~\ref{sec:asat}, and the second term is the routing gradient through the mask selection. Under hard top-$k$ routing, $M_k$ is piecewise constant, so its Jacobian $J_{M_k}$ is zero almost everywhere and undefined at selection ties. We therefore treat the mask as constant during backpropagation without introducing a surrogate gradient. Because the gate projection still receives direct activation gradients through the selected channels, its routing decisions naturally evolve across optimization steps.

Alternative formulations could replace the hard mask during training with a differentiable top-$k$ relaxation~\citep{hou2025instruction} or apply a straight-through estimator (STE) using a surrogate Jacobian in the backward pass. We leave a systematic comparison of these surrogate-gradient variants to future work.

\section{Additional Results}

Unless otherwise stated, Dense denotes the original pretrained checkpoint, and all training-free baselines use the controlled per-token top-$k$ protocol described in Section~\ref{sec:experiments}.

\subsection{Training-Free Gate and Up Routing}

Table~\ref{tab:trainingfree-results} reports the per-task breakdown corresponding to Figure~\ref{fig:gate_vs_up}(a).

\begin{table*}[t]
\centering
\caption{WikiText-2 perplexity and zero-shot accuracy for training-free gate- and up-based routing across model sizes and FFN sparsity levels.}
\label{tab:trainingfree-results}
\begin{adjustbox}{max width=\linewidth}
    \begin{tabular}{l||l|c|ccccccccc}
    
    \toprule\toprule
    
    \textbf{Model} & \textbf{Method} & \textbf{Wiki2$(\downarrow)$}& \textbf{ARC-C} & \textbf{ARC-E} & \textbf{BoolQ} & \textbf{HS} & \textbf{OBQA} & \textbf{PIQA} & \textbf{SIQA} & \textbf{WG} & \textbf{Avg.$(\uparrow)$} \\
    \midrule
    \multirow{9}{*}{LLaMA-3.2-1B} & Dense & 9.7 & 36.4 & 60.4 & 63.9 & 63.6 & 36.8 & 74.5 & 42.9 & 60.3 & 54.9 \\
    & up-25\% & 9.9  & 35.1 & 60.0 & 62.2 & 63.1 & 36.6 & 74.3 & 42.6 & 59.8 & 54.2 \\
    & up-50\% & 11.5  & 33.2 & 57.6 & 58.0 & 60.2 & 37.2 & 72.4 & 41.6 & 57.7 & 52.2 \\
    & up-75\% & 22.8  & 29.3 & 48.4 & 54.3 & 47.3 & 31.4 & 65.6 & 39.3 & 53.4 & 46.1 \\
    & up-85\% & 50.8  & 27.1 & 39.9 & 50.0 & 37.2 & 29.2 & 59.5 & 36.0 & 51.5 & 41.3 \\
    & gate-25\% & 10.1  & 35.8 & 59.7 & 63.0 & 62.4 & 36.8 & 73.6 & 42.5 & 58.6 & 54.1 \\
    & gate-50\% & 15.0  & 33.4 & 54.0 & 59.3 & 55.5 & 32.2 & 69.7 & 40.8 & 55.3 & 50.0 \\
    & gate-75\% & 174.7 & 25.5 & 35.7 & 50.8 & 33.9 & 28.2 & 57.8 & 34.4 & 51.5 & 39.7\\
    & gate-85\% & 17209.7 & 25.9 & 28.3 & 52.8 & 28.3 & 28.8 & 52.6 & 33.5 & 49.6 & 37.5\\
    \midrule
    \multirow{9}{*}{LLaMA-3.2-3B} & Dense & 7.8 & 45.7 & 71.6 & 72.9 & 73.6 & 42.8 & 77.5 & 47.1 & 69.5 & 62.6 \\
    & up-25\% & 7.8  & 46.6 & 71.0 & 72.9 & 73.7 & 42.4 & 77.8 & 47.1 & 69.5 & 62.6 \\
    & up-50\% & 8.5  & 42.3 & 69.2 & 69.4 & 72.3 & 40.4 & 77.9 & 46.2 & 65.9 & 60.5 \\
    & up-75\% & 13.3  & 37.9 & 60.5 & 66.6 & 64.9 & 40.0 & 72.1 & 43.0 & 61.9 & 55.9 \\
    & up-85\% & 25.7  & 32.3 & 49.5 & 54.6 & 52.5 & 30.0 & 66.1 & 40.5 & 52.8 & 47.3 \\
    & gate-25\% & 8.0  & 44.8 & 70.6 & 69.5 & 73.8 & 41.8 & 77.5 & 47.0 & 67.5 & 61.6 \\
    & gate-50\% & 11.2  & 42.2 & 64.7 & 48.8 & 70.3 & 40.6 & 75.8 & 43.4 & 59.7 & 55.7 \\
    & gate-75\% & 128.8 & 30.0 & 40.0 & 47.8 & 45.6 & 31.6 & 60.9 & 35.9 & 52.1 & 43.0\\
    & gate-85\% & 5122.2 & 27.3 & 32.2 & 43.2 & 30.7 & 28.2 & 55.7 & 33.3 & 49.3 & 37.5\\
    
    \midrule
    \multirow{9}{*}{LLaMA-3.1-8B} & Dense & 6.2 & 53.3 & 81.1 & 82.1 & 78.9 & 44.4 & 81.1 & 47.0 & 73.5 & 67.7 \\
    & up-25\% & 6.3  & 53.8 & 80.2 & 81.9 & 78.8 & 45.2 & 81.1 & 47.3 & 73.9 & 67.8 \\
    & up-50\% & 7.2  & 51.5 & 76.8 & 77.2 & 77.8 & 43.4 & 79.5 & 45.9 & 72.0 & 65.5 \\
    & up-75\% & 13.4 & 39.6 & 62.6 & 62.9 & 71.2 & 38.6 & 76.3 & 45.6 & 65.4 & 57.8 \\
    & up-85\% & 27.9  & 34.6 & 53.3 & 48.9 & 60.0 & 35.6 & 70.1 & 41.9 & 58.4 & 50.3 \\
    & gate-25\% & 6.3  & 53.5 & 80.3 & 81.8 & 78.9 & 44.4 & 81.2 & 47.3 & 73.4 & 67.6 \\
    & gate-50\% & 7.8  & 48.8 & 74.3 & 77.9 & 76.8 & 43.0 & 79.8 & 45.0 & 70.1 & 64.5 \\
    & gate-75\% & 28.4 & 34.0 & 50.6 & 55.4 & 63.5 & 37.4 & 72.0 & 40.2 & 54.5 & 51.0\\
    & gate-85\% & 956.8 & 29.0 & 36.9 & 48.8 & 41.6 & 31.0 & 62.1 & 35.3 & 51.5 & 42.0\\

    \bottomrule\bottomrule
    \end{tabular}
  \end{adjustbox}

\end{table*}

\subsection{Quantization Results}
Table~\ref{tab_appendix_w4quant} reports the per-task W$_4$ GPTQ and continued-QAT results across both 75\% and 85\% sparsity, and Table~\ref{tab_qat_detail} provides the per-task dense ParetoQ baseline results used in Figure~\ref{fig:qat_comparison}(a).

\begin{table*}[t]
\centering
\caption{W$_4$ GPTQ and continued-QAT results for checkpoints trained with 75\% or 85\% FFN sparsity. Dense denotes the original pretrained model.}
\begin{adjustbox}{max width=\linewidth}
    \begin{tabular}{l||l|c|ccccccccc}
    
    \toprule\toprule
    
        \textbf{Model} & \textbf{Method} & \textbf{Wiki2$(\downarrow)$}& \textbf{ARC-C} & \textbf{ARC-E} & \textbf{BoolQ} & \textbf{HS} & \textbf{OBQA} & \textbf{PIQA} & \textbf{SIQA} & \textbf{WG} & \textbf{Avg.$(\uparrow)$} \\
    \midrule
    \rowcolor{lightgray}\cellcolor{white}& Dense & 9.7 & 36.4 & 60.4 & 63.9 & 63.6 & 36.8 & 74.5 & 42.9 & 60.3 & 54.9 \\
    \multirow{2}{*}{LLaMA-3.2-1B}  &\textbf{SGA-S}$_{35\text{BT}}$(75\%)+ GPTQ(W$_{4}$) & 13.0  & 38.3 & 66.4 & 59.2 & 58.2 & 39.4 & 73.2 & 41.4 & 57.1 & 54.1 \\
& \cellcolor{sakura!70}\textbf{SGA-S}$_{30\text{BT}}$(75\%)+ QAT$_{5\text{BT}}$(W$_{4}$) & \cellcolor{sakura!70}\textbf{12.6}  & \cellcolor{sakura!70}37.6 & \cellcolor{sakura!70}66.3 & \cellcolor{sakura!70}\textbf{61.9} & \cellcolor{sakura!70}\textbf{58.5} & \cellcolor{sakura!70}\textbf{41.4} & \cellcolor{sakura!70}\textbf{73.2} & \cellcolor{sakura!70}\textbf{41.9} & \cellcolor{sakura!70}\textbf{58.3} & \cellcolor{sakura!70}\textbf{54.9} \\
    &\textbf{SGA-S}$_{35\text{BT}}$(85\%)+ GPTQ(W$_{4}$) & 13.3  & 37.6 & 65.6 & 59.0 & 58.1 & 38.0 & 72.8 & 41.1 & 56.6 & 53.6 \\
\rowcolor{sakura!70}\cellcolor{white}& \textbf{SGA-S}$_{30\text{BT}}$(85\%)+ QAT$_{5\text{BT}}$(W$_{4}$) & \textbf{12.8}  & 37.0 & \textbf{66.5} & \textbf{59.4} & \textbf{58.4} & \textbf{39.8} & 72.3 & 40.5 & \textbf{58.6} & \textbf{54.1} \\
    \midrule
    \midrule
    
     \rowcolor{lightgray}\cellcolor{white}& Dense & 7.8 & 45.7 & 71.6 & 72.9 & 73.6 & 42.8 & 77.5 & 47.1 & 69.5 & 62.6 \\
   \multirow{2}{*}{LLaMA-3.2-3B} &\textbf{SGA-S}$_{35\text{BT}}$(75\%)+ GPTQ(W$_{4}$)& 10.0 & 44.0 & 71.7 & 69.7 & 68.2 & 42.0 & 75.9  & 43.6 & 63.9 & 59.9 \\
    & \cellcolor{sakura!70}\textbf{SGA-S}$_{30\text{BT}}$(75\%)+ QAT$_{5\text{BT}}$(W$_{4}$) & \cellcolor{sakura!70}\textbf{9.8} & \cellcolor{sakura!70}\textbf{48.7} & \cellcolor{sakura!70}\textbf{74.0} & \cellcolor{sakura!70}\textbf{72.1} & \cellcolor{sakura!70}\textbf{68.8} & \cellcolor{sakura!70}\textbf{43.2} & \cellcolor{sakura!70}\textbf{77.5} & \cellcolor{sakura!70}\textbf{44.0} & \cellcolor{sakura!70}\textbf{64.4} & \cellcolor{sakura!70}\textbf{61.6} \\
    &\textbf{SGA-S}$_{35\text{BT}}$(85\%)+ GPTQ(W$_{4}$) & 10.2  & 43.2 & 70.4 & 69.3 & 67.8 & 42.2 & 76.1 & 44.2 & 62.4 & 59.5 \\
\rowcolor{sakura!70}\cellcolor{white}& \textbf{SGA-S}$_{30\text{BT}}$(85\%)+ QAT$_{5\text{BT}}$(W$_{4}$) & \textbf{9.9}  & \textbf{45.7} & \textbf{72.3} & \textbf{71.0} & \textbf{68.1} & \textbf{44.0} & \textbf{76.4} & \textbf{44.5} & \textbf{64.1} & \textbf{60.9} \\
    \midrule
\rowcolor{lightgray}\cellcolor{white}& Dense & 6.2 & 53.3 & 81.1 & 82.1 & 78.9 & 44.4 & 81.1 & 47.0 & 73.5 & 67.7 \\

\multirow{2}{*}{LLaMA-3.1-8B}&\textbf{SGA-S}$_{35\text{BT}}$(75\%)+ GPTQ(W$_{4}$) & 8.2  & 51.4 & 77.1 & 77.6 & 74.9 & 45.4 & 80.0 & 46.4 & 69.1 & 65.2 \\
& \cellcolor{sakura!70}\textbf{SGA-S}$_{30\text{BT}}$(75\%)+ QAT$_{5\text{BT}}$(W$_{4}$) & \cellcolor{sakura!70}\textbf{8.0}  & \cellcolor{sakura!70}\textbf{52.3} & \cellcolor{sakura!70}\textbf{77.4} & \cellcolor{sakura!70}76.1 & \cellcolor{sakura!70}\textbf{75.8} & \cellcolor{sakura!70}\textbf{45.8} & \cellcolor{sakura!70}79.3 & \cellcolor{sakura!70}46.3 & \cellcolor{sakura!70}\textbf{70.8} & \cellcolor{sakura!70}\textbf{65.5} \\

&\textbf{SGA-S}$_{35\text{BT}}$(85\%)+ GPTQ(W$_{4}$) & 8.7  & 50.4 & 77.2 & 73.5 & 73.8 & 44.0 & 77.9 & 45.9 & 66.5 & 63.6 \\
\rowcolor{sakura!70}\cellcolor{white}& \textbf{SGA-S}$_{30\text{BT}}$(85\%)+ QAT$_{5\text{BT}}$(W$_{4}$) & \textbf{8.3}  & \textbf{50.4} & 77.1 & \textbf{75.1} & \textbf{74.8} & \textbf{44.4} & \textbf{80.0} & \textbf{46.8} & \textbf{67.9} & \textbf{64.6} \\

    \bottomrule\bottomrule
    \end{tabular}
  \end{adjustbox}
\label{tab_appendix_w4quant}
\end{table*}

\begin{table*}[t]
\centering
\caption{WikiText-2 perplexity and zero-shot accuracy of the dense ParetoQ
QAT baselines.}
\begin{adjustbox}{max width=\linewidth}
    \begin{tabular}{l||l|c|ccccccccc}
    
    \toprule\toprule
    
    \textbf{Model} & \textbf{Method} & \textbf{Wiki2$(\downarrow)$}& \textbf{ARC-C} & \textbf{ARC-E} & \textbf{BoolQ} & \textbf{HS} & \textbf{OBQA} & \textbf{PIQA} & \textbf{SIQA} & \textbf{WG} & \textbf{Avg.$(\uparrow)$} \\
    \midrule
    \multirow{4}{*}{LLaMA-3.2-1B} & Dense & 9.7 & 36.4 & 60.4 & 63.9 & 63.6 & 36.8 & 74.5 & 42.9 & 60.3 & 54.9 \\
    & ParetoQ(W$_3$) & 11.2  & 37.0 & 65.5 & 63.7 & 61.0 & 37.8 & 73.5 & 42.4 & 59.7 & 55.1 \\
    & ParetoQ(W$_2$) & 11.5  & 35.8 & 64.4 & 61.3 & 57.5 & 39.0 & 73.1 & 41.4 & 58.7 & 53.9 \\
    & ParetoQ(W$_{1.58}$) & 12.0  & 34.0 & 62.7 & 60.6 & 53.0 & 36.6 & 71.1 & 40.8 & 58.1 & 52.1 \\
    \midrule
    \multirow{4}{*}{LLaMA-3.2-3B} & Dense & 7.8 & 45.7 & 71.6 & 72.9 & 73.6 & 42.8 & 77.5 & 47.1 & 69.5 & 62.6 \\
    & ParetoQ(W$_3$) & 8.2   & 46.2 & 73.8 & 71.3 & 71.4 & 41.4 & 77.7 & 45.6 & 67.4 & 61.9 \\
    & ParetoQ(W$_2$) & 8.6  & 43.8 & 72.1 & 70.9 & 69.1 & 43.4 & 77.1 & 44.5 & 65.3 & 60.8 \\
    & ParetoQ(W$_{1.58}$) & 9.3  & 43.2 & 70.3 & 68.0 & 66.2 & 41.0 & 75.3 & 44.9 & 62.1 & 58.9 \\
    
    \midrule
    \multirow{4}{*}{LLaMA-3.1-8B} & Dense & 6.2 & 53.3 & 81.1 & 82.1 & 78.9 & 44.4 & 81.1 & 47.0 & 73.5 & 67.7 \\
    & ParetoQ(W$_3$) & 7.2 & 51.5 & 78.8 & 78.7 & 77.2 & 46.8 & 80.1 & 47.4 & 71.3 & 66.5 \\
    & ParetoQ(W$_2$) & 7.2  & 50.3 & 76.0 & 74.2 & 74.6 & 43.4 & 79.3 & 47.1 & 71.4 & 64.5 \\
    & ParetoQ(W$_{1.58}$) & 8.2  & 50.4 & 75.0 & 65.9 & 72.7 & 43.8 & 77.7 & 47.0 & 69.5 & 62.8 \\

    \bottomrule\bottomrule
    \end{tabular}
  \end{adjustbox}
\label{tab_qat_detail}
\end{table*}

\subsection{Training-Budget Comparisons}

Table~\ref{tab:training-baselines} provides the per-task comparisons across activation-sparsity-aware training baselines under 10B- and 20B-token budgets at 85\% sparsity.

\begin{table*}[t]
\centering
\caption{Per-task results for activation-sparsity-aware training methods under 10B- and 20B-token budgets.}
\begin{adjustbox}{max width=\linewidth}
    \begin{tabular}{c||l|c|ccccccccc}
    
    \toprule\toprule
    
        \textbf{Model Size} & \textbf{Method} & \textbf{Wiki2$(\downarrow)$}& \textbf{ARC-C} & \textbf{ARC-E} & \textbf{BoolQ} & \textbf{HS} & \textbf{OBQA} & \textbf{PIQA} & \textbf{SIQA} & \textbf{WG} & \textbf{Avg.$(\uparrow)$} \\
    \midrule
    \rowcolor{lightgray}\cellcolor{white}& Dense & 9.7 & 36.4 & 60.4 & 63.9 & 63.6 & 36.8 & 74.5 & 42.9 & 60.3 & 54.9 \\
    \cmidrule{2-12}
    & {Q-Sparse}$_{10\text{BT}}$(85\%) & 15.1 & 32.2 & 59.5 & 59.1 & 50.6 & 34.4 & 70.0 & \textbf{41.2} & 55.5 & 50.3 \\
    & {ReLUfication}$_{10\text{BT}}$(85\%) & 18.1 & 32.6 & 57.3 & 60.2 & 56.0 & 36.0 & 70.7 & 39.6 & 53.4 & 50.7 \\
    \rowcolor{sakura!70}\cellcolor{white}& \textbf{SGA-S}$_{10\text{BT}}$(85\%) & \textbf{13.0} & \textbf{35.3} & \textbf{64.2} & \textbf{60.2} & \textbf{56.1} & \textbf{37.8} & \textbf{73.5} & 40.5 & \textbf{58.2} & \textbf{53.2} \\
    \cmidrule{2-12}
    & {Q-Sparse}$_{20\text{BT}}$(85\%) & 14.3 & 33.1 & 61.6 & 60.3 & 52.6 & 35.6 & 71.3 & \textbf{41.0} & \textbf{57.5} & 51.6 \\
    & {ReLUfication}$_{20\text{BT}}$(85\%) & 18.8 & 33.3 & 56.8 & 61.2 & 56.3 & 37.0 & 70.7 & 39.2 & 55.5 & 51.3 \\
    & {TwELL}$_{20\text{BT}}$(85\%) & 15.4 & 34.0 & 63.5 & \textbf{62.3} & 44.6 & 37.2 & 68.7 & 39.2 & 53.2 & 50.3 \\
    \rowcolor{sakura!70}\cellcolor{white}\multirow{-8}{*}{1B}& \textbf{SGA-S}$_{20\text{BT}}$(85\%) & \textbf{12.8} & \textbf{37.3} & \textbf{66.1} & 60.2 & \textbf{58.0} & \textbf{38.0} & \textbf{73.7} & 40.9 & 57.2 & \textbf{53.9} \\
    
    \midrule
    \midrule
    
    \rowcolor{lightgray}\cellcolor{white}& Dense & 7.8 & 45.7 & 71.6 & 72.9 & 73.6 & 42.8 & 77.5 & 47.1 & 69.5 & 62.6 \\
    \cmidrule{2-12}
    & {Q-Sparse}$_{10\text{BT}}$(85\%) & 11.2 & 40.0 & 66.0 & 66.1 & 62.4 & 38.4 & 73.6 & 43.3 & 60.1 & 56.2 \\
    & {ReLUfication}$_{10\text{BT}}$(85\%) & 10.3 & 43.8 & \textbf{71.0} & 69.4 & \textbf{68.2} & 40.6 & \textbf{76.1} & 43.8 & 63.6 & 59.6 \\
    \rowcolor{sakura!70}\cellcolor{white}& \textbf{SGA-S}$_{10\text{BT}}$(85\%) & \textbf{10.1} & \textbf{45.9} & 70.7 & \textbf{70.2} & 67.2 & \textbf{41.2} & \textbf{76.1} & \textbf{45.4} & \textbf{64.8} & \textbf{60.2} \\
    \cmidrule{2-12}
    & {Q-Sparse}$_{20\text{BT}}$(85\%) & 10.9 & 41.7 & 68.8 & 66.1 & 64.1 & 41.4 & 74.8 & 43.6 & 60.1 & 57.6 \\
    & {ReLUfication}$_{20\text{BT}}$(85\%) & 10.2 & 44.2 & 70.7 & 69.5 & \textbf{68.7} & 40.6 & \textbf{76.6} & 44.3 & 63.5 & 59.8 \\
    \rowcolor{sakura!70}\cellcolor{white}\multirow{-7}{*}{3B}& \textbf{SGA-S}$_{20\text{BT}}$(85\%) & \textbf{9.9} & \textbf{46.0} & \textbf{72.5} & \textbf{70.3} & 68.1 & \textbf{44.2} & 76.3 & \textbf{44.6} & \textbf{64.1} & \textbf{60.8} \\

    \bottomrule\bottomrule
    \end{tabular}
  \end{adjustbox}
\label{tab:training-baselines}
\end{table*}

\subsection{Cross-Family Evaluation}
Table~\ref{tab:appendix:qwen} provides the per-task Qwen-3.5-4B results corresponding to Table~\ref{tab:ablation:qwen}.

\begin{table}[t]
\centering
\caption{Per-task results for gate- and up-based routing on Qwen-3.5-4B after 10B training tokens.}
\begin{adjustbox}{max width=\linewidth}
    \begin{tabular}{l||l|cccccccccc}
    
    \toprule\toprule
    
        \textbf{Model} & \textbf{Method} & \textbf{Wiki2$(\downarrow)$}& \textbf{ARC-C} & \textbf{ARC-E} & \textbf{BoolQ} & \textbf{HS} & \textbf{OBQA} & \textbf{PIQA} & \textbf{SIQA} & \textbf{WG} & \textbf{Avg.$(\uparrow)$} \\
    \midrule
     \rowcolor{lightgray}\cellcolor{white}& Dense & 8.0 & 55.2 & 76.3 & 83.9 & 75.3 & 42.4 & 79.5 & 49.2 & 71.1 & 66.6 \\
   & \textbf{SGA-S}$_{10\text{BT}}$(75\%)-up & 9.1  & 55.4 & 79.3 & 76.5 & 71.4 & 43.6 & 77.4 & 45.8 & 68.0 & 64.7 \\
    \multirow{1}{*}{Qwen-3.5-4B}& \cellcolor{sakura!70}\textbf{SGA-S}$_{10\text{BT}}$(75\%)-gate & \cellcolor{sakura!70}\textbf{8.8}  & \cellcolor{sakura!70}\textbf{56.4} & \cellcolor{sakura!70}\textbf{80.6} & \cellcolor{sakura!70}\textbf{78.7} & \cellcolor{sakura!70}\textbf{71.9} & \cellcolor{sakura!70}\textbf{44.2} & \cellcolor{sakura!70}\textbf{77.7} & \cellcolor{sakura!70}\textbf{46.2} & \cellcolor{sakura!70}\textbf{69.2} & \cellcolor{sakura!70}\textbf{65.6} \\
    & \textbf{SGA-S}$_{10\text{BT}}$(85\%)-up & 9.5  & 51.2 & 77.1 & 76.1 & 69.7 & 43.4 & 76.6 & 45.3 & 67.5 & 63.4 \\
    \rowcolor{sakura!70}\cellcolor{white}& \textbf{SGA-S}$_{10\text{BT}}$(85\%)-gate & \textbf{8.9}  & \textbf{55.0} & \textbf{80.0} & \textbf{77.4} & \textbf{71.4} & \textbf{43.4} & \textbf{77.5} & \textbf{45.3} & \textbf{67.6} & \textbf{64.7} \\
    \bottomrule\bottomrule
    \end{tabular}
  \end{adjustbox}
\label{tab:appendix:qwen}
\end{table}

\subsection{Up-Based Routing with 30B Training Tokens}

Table~\ref{tab:appendix:llama_gate_up} reports the per-task results for up-based routing on LLaMA models after 30B training tokens, complementing the gate-based \name results in Table~\ref{tab_main_trainingfree}.

\begin{table}[h]
\centering
\caption{Per-task results for up-based routing on LLaMA models after 30B training tokens.}
\begin{adjustbox}{max width=\linewidth}
    \begin{tabular}{l||l|cccccccccc}
    
    \toprule\toprule
    
    \textbf{Model} & \textbf{Method} & \textbf{Wiki2$(\downarrow)$}& \textbf{ARC-C} & \textbf{ARC-E} & \textbf{BoolQ} & \textbf{HS} & \textbf{OBQA} & \textbf{PIQA} & \textbf{SIQA} & \textbf{WG} & \textbf{Avg.$(\uparrow)$} \\
    \midrule
   \multirow{2}{*}{LLaMA-3.2-1B}& \textbf{SGA-S}$_{30\text{BT}}$(75\%)-up & 12.7  & 38.4 & 67.0 & 63.2 & 57.6 & 39.0 & 73.7 & 40.9 & 56.7 & 54.6 \\
    & \textbf{SGA-S}$_{30\text{BT}}$(85\%)-up & 13.2  & 38.1 & 66.3 & 62.1 & 56.6 & 38.6 & 72.4 & 41.5 & 58.3 & 54.2 \\
    \midrule
   \multirow{2}{*}{LLaMA-3.2-3B}& \textbf{SGA-S}$_{30\text{BT}}$(75\%)-up & 9.9  & 45.6 & 72.5 & 69.3 & 67.9 & 44.2 & 76.4 & 44.5 & 63.9 & 60.5 \\
    & \textbf{SGA-S}$_{30\text{BT}}$(85\%)-up & 10.3  & 43.3 & 71.3 & 69.4 & 67.1
    & 43.2 & 75.8 & 45.4 & 64.4 & 59.8 \\
    \midrule
   \multirow{2}{*}{LLaMA-3.1-8B}& \textbf{SGA-S}$_{30\text{BT}}$(75\%)-up & 8.3  & 49.8 & 75.8 & 76.3 & 74.5 & 44.4 & 79.3 & 47.0 & 69.6 & 64.6 \\
    & \textbf{SGA-S}$_{30\text{BT}}$(85\%)-up & 8.7  & 49.9 & 76.0 & 74.6 & 72.9 & 44.0 & 78.5 & 46.8 & 68.6 & 63.9 \\
    \bottomrule\bottomrule
    \end{tabular}
  \end{adjustbox}
\label{tab:appendix:llama_gate_up}
\end{table}

\subsection{Continued-QAT and Joint-QAT Comparison Results}
Tables~\ref{tab:appendix:lr_cqat} and~\ref{tab:appendix:cqat_vs_jqat} provide the per-task results for continued-QAT (25B sparse training tokens followed by 5B QAT tokens) at both 75\% and 85\% sparsity and joint-QAT (30B tokens) under mixed W$_{2\&4}$ quantization at 75\%, complementing the results in Table~\ref{tab:comparison:qat_methods}.

\begin{table}[h]
\centering
\caption{Per-task results for continued-QAT (25B tokens of sparse training followed by 5B tokens of QAT) at 75\% and 85\% sparsity with W$_{2\&4}$ quantization across LLaMA models.}
\begin{adjustbox}{max width=\linewidth}
    \begin{tabular}{l||l|cccccccccc}
    
    \toprule\toprule
    
    \textbf{Model} & \textbf{Method} & \textbf{Wiki2$(\downarrow)$}& \textbf{ARC-C} & \textbf{ARC-E} & \textbf{BoolQ} & \textbf{HS} & \textbf{OBQA} & \textbf{PIQA} & \textbf{SIQA} & \textbf{WG} & \textbf{Avg.$(\uparrow)$} \\
    \midrule
    \multirow{2}{*}{LLaMA-3.2-1B}& 75\%+W$_{2\&4}$ & 14.8  & 34.9 & 62.5 & 60.2 & 54.3 & 37.8 & 70.6 & 40.4 & 53.9 & 51.8 \\
   & 85\%+W$_{2\&4}$ & 15.1  & 34.4 & 60.3 & 59.2 & 53.7 & 34.8 & 71.9 & 41.4 & 56.8 & 51.6 \\
    \midrule
    \multirow{2}{*}{LLaMA-3.2-3B}& 75\%+W$_{2\&4}$ & 11.1  & 43.7 & 71.6 & 66.1 & 65.7 & 40.2 & 76.0 & 43.5 & 61.3 & 58.5 \\
   & 85\%+W$_{2\&4}$ & 11.2  & 43.7 & 70.8 & 67.1 & 65.1 & 41.2 & 75.7 & 42.9 & 60.1 & 58.3\\
    \midrule
    \multirow{2}{*}{LLaMA-3.1-8B}& 75\%+W$_{2\&4}$ & 9.0  & 49.2 & 75.6 & 72.8 & 73.1 & 46.0 & 78.6 & 44.9 & 68.0 & 63.5 \\
   & 85\%+W$_{2\&4}$ & 9.3  & 49.1 & 75.6 & 72.3 & 72.4 & 44.8 & 78.2 & 45.5 & 66.1 & 63.1 \\
    \bottomrule\bottomrule
    \end{tabular}
  \end{adjustbox}
\label{tab:appendix:lr_cqat}
\end{table}

\begin{table}[h]
\centering
\caption{Per-task results for joint-QAT (30B tokens) at 75\% sparsity with W$_{2\&4}$ quantization across LLaMA models.}
\begin{adjustbox}{max width=\linewidth}
    \begin{tabular}{l||l|cccccccccc}
    
    \toprule\toprule
    
    \textbf{Model} & \textbf{Method} & \textbf{Wiki2$(\downarrow)$}& \textbf{ARC-C} & \textbf{ARC-E} & \textbf{BoolQ} & \textbf{HS} & \textbf{OBQA} & \textbf{PIQA} & \textbf{SIQA} & \textbf{WG} & \textbf{Avg.$(\uparrow)$} \\
    \midrule
   \multirow{1}{*}{LLaMA-3.2-1B}& j-QAT (75\%, W$_{2\&4}$) & 13.4  & 36.6 & 64.1 & 60.1 & 56.4 & 35.6 & 73.5 & 41.2 & 56.9 & 53.1 \\
    \midrule
   \multirow{1}{*}{LLaMA-3.2-3B}& j-QAT (75\%, W$_{2\&4}$) & 10.3  & 45.3 & 72.8 & 67.6 & 67.0 & 43.6 & 75.7 & 44.6 & 62.4 & 59.9 \\
    \midrule
   \multirow{1}{*}{LLaMA-3.1-8B}& j-QAT (75\%, W$_{2\&4}$) & 8.6  & 51.3 & 76.7 & 73.7 & 74.1 & 44.8 & 80.0 & 45.9 & 70.0 & 64.6 \\
    \bottomrule\bottomrule
    \end{tabular}
  \end{adjustbox}
\label{tab:appendix:cqat_vs_jqat}
\end{table}

\end{document}